\documentclass[11pt]{article}

\usepackage[final]{acl}

 \usepackage{microtype}

\usepackage{booktabs}
\usepackage{ragged2e}
\usepackage{amsmath, amsthm, amssymb, amsfonts}
\usepackage{tabularx}
\usepackage[autostyle=false, style=english]{csquotes}
\MakeOuterQuote{"}

\newtheorem{problem}{Problem}
\newcommand{\problemname}[1]{\textsc{#1}} 

\title{NP-Hard Lower Bound Complexity for Semantic Self-Verification}

\author{Robin Young \\
  Department of Computer Science and Technology \\
  University of Cambridge \\
  Cambridge, UK \\
  \texttt{robin.young@cl.cam.ac.uk}}

\begin{document}

\maketitle
\begin{abstract}
We model Semantic Self-Verification (SSV) as the problem of determining whether a statement accurately characterizes its own semantic properties within a given interpretive framework that formalizes a challenge in AI safety and fairness: can an AI system verify that it has correctly interpreted rules intended to govern its behavior? We prove that SSV, in this specification, is NP-complete by constructing a polynomial-time reduction from 3-Satisfiability (3-SAT). Our reduction maps a 3-SAT formula to an instance of SSV involving ambiguous terms with binary interpretations and semantic constraints derived from logical clauses. This establishes that even simplified forms of semantic self-verification should face computational barriers. The NP-complete lower bound has implications for AI safety and fairness approaches that rely on semantic interpretation of instructions, including but not limited to constitutional AI, alignment via natural language, and instruction-following systems. Approaches where an AI system verify its understanding of directives may face this computational barrier. We argue that more realistic verification scenarios likely face even greater complexity.
\end{abstract}

\section{Introduction}
\label{sec:introduction}

As artificial intelligence systems become increasingly sophisticated, ensuring they understand and correctly interpret their directives becomes a critical challenge. How can it verify that it has correctly interpreted these directives? This question of semantic self-verification lies at the heart of many AI safety and fairness problems.

In this paper, we propose and formalize Semantic Self-Verification (SSV) as the computational problem of determining whether a statement accurately characterizes its own semantic properties within a given interpretive framework. This formalization captures a challenge: can a system verify the correctness of its own understanding?

AI safety faces related self-reference challenges \cite{ji2025aialignmentcomprehensivesurvey}. Constitutional AI approaches guide system behavior through principles the system interprets and applies. However, the system itself determines whether its interpretations of these principles are correct, which is a self-referential task. Prior work on mechanistic interpretability and automated verification in AI systems has attempted to address aspects of this challenge \cite{Katz2017ReluplexAE}, but the computational complexity of semantic self-verification has remained undercharacterized.




We demonstrate that Semantic Self-Verification is at least NP-hard by constructing a polynomial-time reduction from \textsc{3-Satisfiability} (3-SAT). This establishes that even simplified versions of semantic self-verification face significant computational barriers. Specifically, we show that determining whether there exists a consistent interpretation of ambiguous terms that satisfies a set of semantic constraints is at least as hard as solving NP-complete problems.


The lower bound we establish provides a corollary heuristic for evaluating safety or fairness system claims. If a system responds too quickly when facing complex semantic verification challenges, it cannot possibly be performing complete verification: the computational lower bound establishes minimum time requirements for the worst case.

This suggests that practical semantic instruction implementations necessarily employ approximations rather than complete verification, raising questions about the reliability and upper bound of guarantees and highlighting the need for transparency about verification limitations. The contribution here is a formal proof whose utility is in establishing a computational lower bound, thereby providing a lens for the design and evaluation of systems that rely on semantic self-verification.

\section{Related Work}
\label{sec:related}

\subsection{Computational Complexity in AI}

The application of computational complexity theory to AI alignment problems has been a valuable framework for understanding limitations. Computational complexity characterizes problems by the resources required to solve them, providing an independent measure of intrinsic difficulty.

Previous work \cite{Wojtowicz2024} has demonstrated that informational persuasion is NP-complete, establishing that discovering persuasive arguments is computationally hard while verifying them remains tractable. This asymmetry helps explain why humans are susceptible to persuasion even when all relevant information is publicly available. Our work on Semantic Self-Verification focuses instead on the computational difficulties systems face when verifying their own semantic interpretations. Complexity theory has also demonstrated the computational complexity of logical induction in alignment contexts \cite{demski2020embeddedagency}. These results suggest that many alignment challenges face not just engineering difficulties but computational barriers.

\subsection{Constitutional AI and Self-Verification}

Constitutional AI approaches aim to align AI systems by providing them with principles they should follow \cite{Bai2022ConstitutionalAH}. These approaches typically rely on the system's ability to interpret these principles. However, the computational complexity of this verification process has not been previously characterized. Practical methods cleverly sidestep a direct formal approach; instead of verifying outputs against a constitution, they distill its principles into a heuristic reward signal for reinforcement learning. This raises a question. What computational barrier makes this indirect, heuristic approach a necessity?

The constitutional approach relates to earlier work on value alignment through principles \cite{Russell2019HumanCA, HadfieldMenell2016CooperativeIR} and rule-based ethical frameworks \cite{Anderson2011-SANME}. Similar challenges arise in other instruction-following paradigms \cite{Ouyang2022TrainingLM}, including reinforcement learning from human feedback (RLHF) \cite{Christiano2017DeepRL}.

Our work provides a complexity-theoretic perspective on these approaches, suggesting there exist fundamental barriers to systems fully verifying their alignment with constitutional principles or other semantic directives.





\subsection{Formal Verification}

Formal verification aims to provide guarantees about system behavior. Existing work has applied these techniques to neural networks and other systems \cite{Katz2017ReluplexAE, seshia2016formal, Cimatti2018IncrementalLF, Sidrane2022}. These approaches typically focus on verifying specific properties rather than semantic understanding.

The computational complexity of verification has been explored in specific contexts of neural networks. For example, previous work \cite{Katz2017ReluplexAE} showed that verifying ReLU-activated neural networks alone is NP-complete and established complexity results for robustness verification \cite{wang2022efficientglobalrobustnesscertification, Liu2019AlgorithmsFV}.

Our work examines instead the complexity of semantic verification specifically, addressing the challenge of verifying that a system has correctly interpreted meaning rather than exhibiting correct input-output behavior.

\subsection{Semantic Understanding in AI Systems}

Recent advances in large language models have focused attention on semantic understanding capabilities and limitations \cite{Bender2020ClimbingTN, ji2025aialignmentcomprehensivesurvey}. Work on mechanistic interpretability \cite{olah2018the, olah2020zoom, elhage2021mathematical} attempts to understand how these models represent and process meaning.

The challenge of verifying semantic understanding relates to debates about symbol grounding \cite{Harnad1990} and the capabilities and limitations of statistical pattern matching vs. understanding \cite{Mitchell2021}. Many are well familiar with the language models are stochastic parrots argument \cite{Bender2021}, while others contend that sufficiently advanced models may develop forms of understanding \cite{bubeck2023sparksartificialgeneralintelligence}.
Our complexity result provides a computational perspective on these debates, suggesting that regardless of whether systems can understand meaning, verifying this understanding faces computational barriers.

\section{The \textsc{3-Satisfiability} Problem (3-SAT)}
\label{sec:3sat}

To understand the difficulty of our Semantic Self-Verification problem, we compare it to the canonical problem of \textsc{3-Satisfiability} \cite{Cook1971TheCO, Garey1979JohnsonCA}. If we can show that solving 3-SAT is a special case of solving SSV, then SSV must be at least as hard as 3-SAT.

\begin{problem}[\problemname{3-SAT}]
    \leavevmode
    \begin{itemize}
        \item \textbf{Input:} A Boolean formula \(\phi\) in 3-conjunctive normal form (3-CNF). A 3-CNF formula is a conjunction (AND) of clauses \(C_1 \land C_2 \land \dots \land C_m\), where each clause \(C_j\) is a disjunction (OR) of exactly three literals \(l_{j1} \lor l_{j2} \lor l_{j3}\). A literal is a Boolean variable \(x_i\) or its negation \(\neg x_i\).
        \item \textbf{Output:} TRUE if there exists a truth assignment (a way of setting variables to TRUE or FALSE) for the variables \(x_1, \dots, x_n\) that makes the entire formula \(\phi\) true; FALSE otherwise.
    \end{itemize}
\end{problem}

\section{Formal Definition of Semantic Self-Verification (SSV)}
\label{sec:ssv_definition}

We now formally define the core problem of Semantic Self-Verification (SSV). Intuitively, SSV asks whether a given statement is telling the truth about its own meaning, according to a specific set of rules for interpreting it. This formalization aims to capture situations where an AI might need to check if its understanding of a rule or principle is internally consistent and valid according to how terms are defined and constrained. For the purposes of formal analysis, we will model semantic interpretation using a simplified framework of discrete choices and logical constraints, aiming to capture a core combinatorial challenge inherent in even restricted forms of self-verification.

\subsection{Problem Statement (Decision Problem)}
\label{ssec:ssv_problem_statement}

\begin{problem}[\problemname{Semantic Self-Verification (SSV)}]
    \leavevmode
    \begin{itemize}
        \item \textbf{Input:}
            \begin{enumerate}
                \item A statement \(S\).
                \item A semantic framework \(F\).
            \end{enumerate}
        \item \textbf{Output:} TRUE if statement \(S\) accurately describes its own semantic properties under framework \(F\); FALSE otherwise.
    \end{itemize}
\end{problem}

\subsection{Components of the Semantic Framework {\(F\)}}
\label{ssec:ssv_framework_components}

To make SSV precise, we define what a "semantic framework" consists of. This framework provides the necessary components for interpreting the statement \(S\) and verifying its claims about itself.

For the purpose of our reduction, we define the semantic framework \(F = (T, \Sigma, \text{Cons}, M, V)\) where:
\begin{itemize}
    \item \(T = \{t_1, t_2, \dots, t_n\}\): A finite set of \textit{ambiguous terms} present in or implicitly referred to by statement \(S\).
    
    These are words or phrases in statement \(S\) that could have multiple meanings.
    
    \item \(\Sigma = \{\sigma_{1}^0, \sigma_{1}^1, \sigma_{2}^0, \sigma_{2}^1, \dots, \sigma_{n}^0, \sigma_{n}^1\}\): A finite set of possible \textit{elemental meanings} or \textit{senses}. In our simple model, each ambiguous term \(t_i \in T\) is associated with exactly two possible elemental meanings: \(\sigma_i^0\) and \(\sigma_i^1\).
    
    For each ambiguous term \(t_i\), we give it exactly two possible interpretations, \(\sigma_i^0\) or \(\sigma_i^1\). This binary choice will directly correspond to the TRUE/FALSE assignments in 3-SAT.
    
    \item \(\text{Cons} = \{SC_1, SC_2, \dots, SC_m\}\): A finite set of \textit{semantic constraints}. Each constraint \(SC_j\) is a condition on the meanings assigned to a subset of terms in \(T\).
    
    These are rules that dictate which combinations of meanings are allowed. For example, a constraint might say "If term \(t_1\) has meaning \(\sigma_1^0\), then term \(t_2\) cannot have meaning \(\sigma_2^1\)."
    
    \item \(M\): The \textit{Meaning Function}.
        \begin{itemize}
            \item An \textit{interpretation} \(I\) of statement \(S\) (with respect to its ambiguous terms in \(T\)) is a function \(I: T \to \bigcup_{i=1}^{n} \{\sigma_i^0, \sigma_i^1\}\) such that for each \(t_i \in T\), \(I(t_i) \in \{\sigma_i^0, \sigma_i^1\}\).
            
            An interpretation is simply a choice of one specific meaning (out of its two possibilities) for every ambiguous term in the statement.
            
            \item \(M(S, F)\) (or simply \(M(S)\) when F is clear) is the set of all \(2^n\) possible interpretations of \(S\).
            
            If there are \(n\) ambiguous terms, each with 2 meanings, there are \(2^n\) ways to interpret the statement as a whole.
            
        \end{itemize}
    \item \(V\): The \textit{Verification Function}.
        \begin{itemize}
            \item \(V(S, I, F)\) (or \(V(S, I)\)) = TRUE if interpretation \(I\) of \(S\) satisfies all semantic constraints in \(\text{Cons}\); FALSE otherwise.
            
            This function checks if a particular interpretation \(I\) is "valid" by seeing if it obeys all the rules (semantic constraints).
            
            \item The overall SSV problem then asks whether \(S\)'s specific self-referential claim about the existence of such a satisfying interpretation is true.
            
            Statement \(S\) itself will make a claim (e.g. "There exists an interpretation of my terms that satisfies all my constraints"). The SSV problem is about checking if that very claim made by \(S\) is correct according to the framework \(F\).
        \end{itemize}
\end{itemize}

\section{Intuitive Example of SSV}

Consider an AI system with constitutional principles: "be helpful," "be honest," and "protect privacy." Each term has binary interpretations: "helpful" means either "provide what users request" ($\sigma^0_1$) or "serve users' best interests" ($\sigma^1_1$); "honest" means "don't state falsehoods" ($\sigma^0_2$) or "actively correct misconceptions" ($\sigma^1_2$); "privacy" means "don't share data externally" ($\sigma^0_3$) or "don't record conversations" ($\sigma^1_3$).

These interpretations face logical constraints. If helpful = "provide requests" ($\sigma^0_1$), then privacy $\neq$ "don't record" ($\sigma^1_3$), since effective responses require conversation history. If honest = "correct misconceptions" ($\sigma^1_2$), then helpful must = "serve best interests" ($\sigma^1_1$), since corrections may contradict explicit requests.

The semantic self-verification question becomes: "Does there exist a consistent interpretation satisfying all constraints?" With 3 terms and 2 interpretations each, there are $2^3 = 8$ possible combinations to check. For $n$ terms, this grows to $2^n$ combinations, each requiring constraint verification, thus creating the combinatorial explosion that makes SSV computationally hard.

This captures a core challenge: verifying that safety directives have coherent, mutually consistent interpretations. Our NP-hardness result formalizes why this verification problem is fundamentally intractable, even in simplified form.

\section{Proof Sketch of NP-Completeness}

We establish our main result by proving that Semantic Self-Verification (SSV) is both NP-hard and a member of the class NP. A full proof is provided in Appendix~\ref{sec:np_hardness_proof}.

\subsection{NP-Hardness Reduction from 3-SAT}

We demonstrate NP-hardness via a polynomial-time reduction from 3-SAT. We construct an SSV instance $(S_\phi, F_\phi)$ from an arbitrary 3-SAT formula $\phi$ such that the SSV instance is TRUE if and only if $\phi$ is satisfiable.

The mapping is as follows:
\begin{itemize}
    \item \textbf{Variables to Terms:} Each Boolean variable $x_i$ in $\phi$ is mapped to a unique ambiguous term $t_i$ in the SSV framework.
    \item \textbf{Truth Values to Meanings:} The TRUE/FALSE assignment of a variable $x_i$ is mapped to a binary choice of interpretation for its corresponding term $t_i$. For example, $I(t_i) = \sigma_i^1$ can represent $x_i=\text{TRUE}$ and $I(t_i) = \sigma_i^0$ can represent $x_i=\text{FALSE}$.
    \item \textbf{Clauses to Constraints:} Each logical clause $C_j$ in $\phi$ (e.g. $x_1 \lor \neg x_2 \lor x_3$) is converted into a semantic constraint $SC_j$. This constraint is satisfied by an interpretation $I$ if and only if the chosen meanings for its terms would make the original clause $C_j$ true.
\end{itemize}

The core of the reduction lies in the specific self-referential statement, $S_\phi$. This statement is constructed to make the claim: \textit{"There exists at least one interpretation of my terms that simultaneously satisfies all of my semantic constraints."} The SSV problem then asks whether this statement is true. By our construction, this question is logically equivalent to asking whether there exists a satisfying truth assignment for the original 3-SAT formula $\phi$. The transformation from $\phi$ to $(S_\phi, F_\phi)$ is efficient, requiring only linear time in the size of the formula.

\subsection{Membership in NP Verification}

To prove that SSV is in NP, we show that a "yes" answer has a certificate that can be verified in polynomial time. For an SSV instance that is TRUE, the certificate is simply a specific interpretation $I^*$ that satisfies all semantic constraints.

A verification algorithm can check this certificate efficiently:
\begin{enumerate}
    \item Given the certificate $I^*$, iterate through each of the $m$ semantic constraints in the framework $F_\phi$.
    \item For each constraint, check if the meanings assigned by $I^*$ to its terms satisfy the constraint. This is a constant-time operation for each constraint.
\end{enumerate}

The total verification time is therefore polynomial in the size of the input instance ($O(m+n)$), confirming that SSV is in NP. Since SSV is both NP-hard and in NP, it is NP-complete.

\section{Discussions}
\label{sec:discussions}

The NP-completeness of SSV demonstrates a computational barrier inherent in systems attempting to verify their own semantic understanding. While our formalization abstracts from the full complexities of natural language that have richer ambiguities, nuanced semantic constraints beyond logical clauses, and pervasive context-dependency, the NP-hardness result for this core problem is suggestive that the difficulty is not merely an artifact of highly expressive semantics but is present even when the interpretive space is highly constrained.

The difficulty stems from two primary sources: first, the exponential growth in possible interpretations ($2^n$ in our binary case); and second, the combinatorial challenge of finding globally consistent interpretations when the choice of meaning for one term constrains valid choices for others. This provides a computational complexity lens that posits that achieving provably correct semantic self-understanding is constrained not only by what can be known or specified, but by what can be feasibly computed.

Recognizing computational limitations, researchers have proposed various pragmatic approaches. These include corrigibility measures that allow systems to be corrected \cite{Soares2017AgentFF, Carey2023}, impact measures that limit potential harm \cite{Krakovna2018PenalizingSE}, uncertainty measurements \cite{hendrycks2021aligning, Hendrycks2019}, and approaches that use human feedback \cite{Stiennon2020}.

Out of that, RLHF has emerged as the dominant alignment paradigm, with seminal work \cite{Christiano2017DeepRL} establishing the approach and subsequent refinements by OpenAI \cite{Ouyang2022TrainingLM} and Anthropic \cite{Bai2022TrainingAH}. Constitutional AI, proposed by Anthropic \cite{Bai2022ConstitutionalAH}, extends RLHF by having systems critique their own outputs according to constitutional principles, with recent work exploring red-teaming \cite{Perez2022RedTL} and recursive evaluation approaches \cite{Lee2024RLAIF}.

Other practical approaches include AI safety via debate \cite{Irving2018AISV, brown-cohen2024scalable}, where systems justify their reasoning through structured argumentation, and process-based supervision \cite{Bowman2022MeasuringPO} that focuses on how systems reach conclusions rather than just the conclusions themselves. Chain-of-thought prompting \cite{wei2022chain} and self-consistency techniques \cite{wang2023selfconsistency} represent lighter-weight mechanisms to improve reasoning without full verification.

Our complexity result complements the importance of these pragmatic approaches by establishing that perfect self-verification may be fundamentally intractable, necessitating approximations and external verification mechanisms. This computational barrier suggests that approaches relying on approximate alignment through iterative refinement may be not just practically useful but theoretically necessary given the intractability of perfect self-verification. Yet, it also reminds that this reliance on approximation creates an inherent gap between intended behavior and achievable guarantees, a gap whose significance may grow with increasing system capability and autonomy.

Our result has several implications for AI safety approaches that rely implicitly or explicitly on self-verification. AI systems tasked with adhering to safety protocols, ethical guidelines, or constitutional principles that interpret these directives and verify their compliance. Our result indicates that performing such verification comprehensively is computationally intractable in general.

This impacts the strong interpretation of constitutional AI or formal verification approaches. The idea that an AI can reliably and exhaustively verify its compliance with a complex constitution faces this barrier. The act of ensuring interpretation of constitutional articles is internally consistent and satisfies all inter-article dependencies mirrors an SSV problem. The result demonstrates that these methods are not, and cannot be, exercises in verified guarantees. Instead, they should be understood as heuristic methods for navigating an intractable search space. 

The result supports a response-time heuristic for evaluating alignment claims. Given that NP-hard problems typically require exponential time in the worst case, if an AI system provides responses involving complex semantic interpretation within human-acceptable latencies, it cannot be performing complete SSV. This implies verification processes are necessarily approximations rather than verification. While such approximations are practically effective, the potential for unverified edge cases remains a concern when guarantees of complete adherence are sought.

While the response-time heuristic serves as a useful, high-level critique against claims of verification at inference, a more interesting application of our SSV result applies to the training process of modern alignment techniques like Constitutional AI. In the RLAIF pipeline, an AI model is tasked with critiquing and rewriting responses. We posit that this critique-and-rewrite cycle is a practical, high-dimensional instance of the SSV problem we have formalized. The AI reviewer interprets the meanings of multiple constitutional articles, check for violations, and generate an output that constitutes a satisfying interpretation. Given that this task is at least NP-complete, this suggests the AI generating this preference data is not performing a complete, logically exhaustive verification. It may be, by computational necessity, using learned heuristics to find a plausible solution, not a correct one.

The preference data used to align the final model is itself the product of a computationally bounded, approximate process. We hypothesize the RLAIF method can be seen as an approximation of an approximation: the final model learns a statistical representation of preference data which was a heuristic guess at what compliance with the constitution actually means. The guarantee of alignment is thus no stronger than the heuristic capabilities of the model that created the training data, and our result demonstrates this is a foundation built on potentially exploitable, computationally necessary compromise. This process distills the web of semantic constraints into a scalar reward signal, which a policy then learns to maximize.

This maneuver, while pragmatic, trades the known difficulty of formal intractability for the arguably more perilous problem of statistical fidelity in a high-dimensional, (mostly) non-interpretable latent space. Instead of satisfying explicit logical constraints, the model learns to align with the statistical shape of a preference model. The persistent phenomenon of jailbreaking in such systems could therefore be seen as an analogous empirical manifestation of our complexity result. These jailbreaks are the adversarial, worst-case instances that an approximate, heuristic method is least equipped to handle. They represent the system finding a path through latent space that maximizes its reward signal while simultaneously violating the core semantic intent of its constitution. This suggests a fundamental ceiling on the worst-case reliability of a strategy that relies on such an approximation, a ceiling that becomes more concerning as a model's capabilities grow.

Our findings also highlight scaling challenges. As rule sets become more extensive or semantically richer, corresponding SSV instances grow in complexity, exacerbating computational burden. This suggests that merely adding more rules to an AI's constitution might not straightforwardly improve alignment if the system cannot feasibly verify adherence to the augmented set.

This computational limit, in our view, could be an informative design constraint. It pushes practitioners from the Sisyphean task of perfect, real-time verification toward the more subtle engineering challenge of intelligent approximation. This suggests several avenues for future work and immediate practice. The core challenge is not just to be approximate, but to be approximate in ways that fail gracefully and predictably.

One such avenue could be related to rule design. Instead of adding more rules to a constitution, designers should consider their structure. The hardness of this problem stems from the dense web of interdependencies between interpretations. Therefore, a more robust constitution might be one composed of modular principles with minimal semantic overlap, effectively aiming to (as far as that is possible) decouple the constraint graph and containing the combinatorial explosion. Structuring directives to resemble computationally tractable logical forms (like, as an analogy, Horn clauses) rather than tangled, arbitrary clauses could turn an intractable general problem into a series of solvable special cases. The goal is to build a constitution that is both ethically sound and also computationally tractable.

Conceptually, the difference is one of global search versus logical propagation. A constitution with broad disjunctive clauses (k-SAT) presents a verifier with a knot. ("Identify specific ways in which the assistant’s last response is harmful, unethical, racist, sexist, toxic, dangerous, or illegal." \cite{Bai2022ConstitutionalAH} is something analogous to 7-SAT.) To check compliance, a hypothetical system holds all constraints in its head at once, searching the entire combinatorial space for a single, consistent interpretation. It is a holistic puzzle where the validity of one choice cannot be known without knowing all the others. A tractable constitution, by contrast, may behave more like a chain of falling dominoes (as in Horn clauses). Verification begins with the given context and unit propagates forward: if this premise is true, then that rule is in effect, setting a specific interpretation which, in turn, may trigger the next. The process is a linear cascade, not an exponential search. To be clear, this analogy to Horn clauses and a tractable constitution is structural, not implementational. The proposal is not a literal call to build systems from vast, brittle trees of if/else statements, a paradigm whose limitations are well-known.

We do not suggest that an LLM behaves like a HORNSAT solver or we should make it behave like one. Rather, this distinction clarifies the nature of the value learning task itself. Forcing a model to learn from preferences governed by a tangled constitution is asking it to approximate a solution to an intractable problem; we should expect its learned heuristics to be brittle and fail on adversarial edge cases. The goal, perhaps, could be to design systems whose reasoning pathways are, by their nature, tractable cascades rather than intractable holistic searches. The "clauses" of such a constitution might be implemented by dedicated, specialized models; the "propagation" might be a structured chain of prompts or a modular composition of functions. The takeaway is to impose a tractable, hierarchical logic on top of the powerful but unstructured capabilities of the underlying models, thereby making the system's reasoning more predictable and verifiable, especially in worst-case scenarios. Training it on preferences governed by a tractable constitution is asking it to learn a simpler underlying logical structure of the policy and thus making predictable reasoning the path of least statistical resistance.

\section{Future Work}

The work in this paper has focused on establishing the computational complexity of what can be termed the \textit{Existence Problem} for SSV: determining if \textit{at least one} internally consistent interpretation of a set of directives exists. We proved this problem is NP-complete, establishing a lower bound on the difficulty of verifying a semantic framework.

From an AI safety perspective, however, a more stringent and often more useful question arises. A system might find a valid interpretation of its rules, but if other, equally valid interpretations exist that lead to contrary behaviors, the system faces a concerning ambiguity. This motivates a subsequent, harder problem.

We can define the \textsc{Semantic Uniqueness} (SU) problem as follows:

\begin{itemize}
    \item \textbf{Input:} Similar to SSV, a statement $S$, a semantic framework $F=(T, \Sigma, \text{Cons}, M, V)$, and a subset of safety-critical terms $\tau \subseteq T$.
    \item \textbf{Output:} TRUE if all valid interpretations of $S$ under $F$ agree on the meaning assigned to every term in $\tau$; FALSE otherwise.
\end{itemize}

Formally, let $\mathcal{I}_{\text{valid}} = \{I \in M(S,F) \mid V(S,I,F) = \text{TRUE}\}$ be the set of all valid interpretations. The SU problem asks if it is true that:
\[
    \forall I_1, I_2 \in \mathcal{I}_{\text{valid}}, \; \forall t \in \tau, \; I_1(t) = I_2(t)
\]

Here we can see a natural extension of our model, which mirrors the distinction between SAT and UNIQUE-SAT \cite{Valiant1985NPIA}.

At a glance, the logical structure of verifying a property that holds for all valid interpretations suggests it belongs to the class co-NP. Its complement, which we can call Ambiguity (i.e. "do there exist two valid interpretations that disagree on $\tau$?"), is readily seen to be in NP by intuition. A certificate for Ambiguity is just the pair of interpretations $(I_1, I_2)$ that conflict; we can verify in polynomial time that both are valid but assign different meanings to a term in $\tau$.

However, further analysis reveals a logical predicate hidden within the question of uniqueness: the set of valid interpretations has to be non-empty. A unique solution must both exist and be the only one. Therefore, the SU problem is more accurately characterized as the conjunction of two distinct conditions:
\begin{itemize}
    \item \textbf{Existence:} There exists at least one valid interpretation $I$ satisfying all semantic constraints. This is the Semantic Self-Verification (SSV) problem, which we have proven to be NP-complete.
    \item \textbf{Uniqueness:} There does not exist a pair of valid interpretations $(I_1, I_2)$ that disagree on the meaning of any term in the set $\tau$. As argued above, this is a co-NP property.
\end{itemize}

A problem defined by the intersection of a language in NP and a language in co-NP belongs to DP complexity (Difference Polynomial-Time) \cite{Papadimitriou1982TheCO}. The canonical DP-complete problem, \textsc{UNIQUE-SAT}, asks if a given Boolean formula has exactly one satisfying assignment, mirroring the structure of SU.

\textbf{Conjecture.} \textit{The Semantic Uniqueness (SU) problem is DP-complete.}

This distinction is important to robust AI safety. If this conjecture holds, it implies that ensuring a directive has a single, unambiguous interpretation is likely strictly harder than merely ensuring it has at least one. A FALSE output for the SU problem is a formal warning of a semantic loophole: the AI has identified multiple, mutually incompatible but equally valid ways to interpret its rules. This is the point at which an aligned system should halt or query a human operator for clarification, rather than proceeding by arbitrarily selecting an interpretation that, while valid, may not be the one intended. Characterizing the hardness of this problem is a natural next step in mapping the computational complexity properties of AI safety.

\section{Conclusion}
\label{sec:conclusion}

We established that Semantic Self-Verification (SSV) is NP-complete, even under a simplified formulation with binary interpretations and clause-based constraints. This result reveals a computational barrier where the combinatorial explosion of possible meanings and their interdependencies creates a verification problem that cannot be efficiently solved in the worst case.

The NP-hardness lower bound has implications for AI alignment approaches that rely, implicitly or explicitly, on semantic self-verification. A corollary of the results suggests that systems responding quickly to complex constitutional constraints cannot be performing complete verification, providing a practical response-time heuristic for evaluating alignment claims. 

While the ultimate goal of a safety system may be to ensure a unique, intended interpretation, we first address the logically prior problem. By proving that even this simpler task is NP-complete, we establish a lower bound on the computational complexity of any more stringent verification.

Rather than cause for pessimism, this result points toward productive research directions. It highlights the necessity of approximation strategies, the importance of designing constitutional frameworks with computational complexity in mind, and the value of hybrid approaches combining limited verification with external oversight. Understanding which approximations preserve safety properties becomes a useful research frontier.

We hope our complexity results help distinguish tractable approaches from those that face fundamental computational barriers. Just as cryptography has advanced by embracing computational hardness results, AI safety research can benefit from acknowledging these theoretical constraints while developing practical methods that work within them.

\section{Limitations}

Our NP-completeness proof involves simplifying ambiguous terms to binary interpretations. One might argue this oversimplifies real-world semantics where terms often possess multiple discrete meanings or continuous semantic spaces. However, this simplification does not undermine the lower bound. Any set of $k$ discrete choices can be represented using $\lceil \log_2 k \rceil$ binary variables. If semantic constraints over $k$-ary choices can be translated into equivalent constraints over binary encodings while preserving the core structure of finding a satisfying assignment, the problem remains in NP and the lower bound holds. Thus, intractability is not merely an artifact of binary choice but is inherent in the combinatorial search for consistent assignments across interdependent, constrained choices.

Any attempt to formalize a continuous space would either quantize the space into a finite set of discrete bins, or employ a formalism that can reason over real numbers. In the first case, achieving meaningful precision would require a large number of bins, making the resulting discrete problem larger, while still being subject to the NP-complete lower bound. In the second case, the problem shifts from the realm of propositional satisfiability to more demanding frameworks like Satisfiability Modulo Theories (SMT).

Therefore, whether semantics are modeled as constrained $k$-ary discrete choices or as continuous scales, the conclusion remains the same. Our NP-completeness result for the simplest discrete case establishes a computational floor. We conjecture that more realistic SSV models, accounting for richer semantic structures, likely face even greater computational complexity, potentially residing in PSPACE or higher. Consider scenarios involving:

\paragraph{Recursive Semantics and Context-Dependency} The interpretation of one rule might depend on another's interpretation within a nested context (e.g. "interpret rule A, then interpret rule B which refers back to how A was handled"). This recursive structure, where recursion depth could be polynomial in input size, is characteristic of PSPACE.

\paragraph{Quantified Constraints} If terms have $n$-ary choices and semantic constraints involve quantified conditions ("for any interpretation of term $t_1$, there must exist an interpretation of term $t_2$ such that..."), this mirrors Quantified Boolean Formulas (QBF), a canonical PSPACE-complete problem.

One might object by appealing to average-case complexity, hoping that practical SSV instances are tractable. This hope is misplaced in a safety context. Safety is determined by worst-case resilience, not average-case performance. An adversary, whether a human red team or the system's own motivated reasoning, does not generate typical problems; it searches for the worst-case instances that cause verification to fail. The practice of jailbreaking language models with convoluted prompts \cite{Perez2022RedTL} is a real-world search for these computationally hard semantic edge cases.

Then, we might also question whether the explicit, logic-based formalization of semantic constraints reflects how meaning operates in real systems, where constraints emerge implicitly from learning patterns. However, this criticism overlooks how semantic constraints genuinely operate in natural language. Consider, as a famous example \cite{Chomsky57a}, "colorless green ideas sleep furiously." While grammatically correct, it violates multiple semantic constraints that any competent language user intuitively recognizes: contradictory properties ("colorless" yet "green"), category errors (abstract "ideas" with physical properties), and action incompatibilities ("sleep furiously"). 

These violations demonstrate that meaningful interpretation in natural language is already governed by constraints, even if not explicitly represented as logical rules. Our SSV formalization doesn't introduce something foreign to semantics; it makes explicit what exists implicitly for necessary analytical purposes. While real semantic systems may represent these constraints in distributed, probabilistic patterns, the essential combinatorial structure of semantic constraint satisfaction remains. If even our simplified formalization leads to NP-completeness, the more complex constraint satisfaction problem in real systems is unlikely to be computationally simpler.

While formal proof of PSPACE-hardness for enriched SSV models is beyond our scope here, the NP-complete lower bound for our simplified model serves as a foundation. It establishes that even before adding richer semantic complexity which intuitively should only make the problem harder, we already encounter computational intractability. Our result therefore provides a robust floor for understanding limits in more expressive semantic models.

\bibliography{custom}

\appendix

\section{Appendix: NP-Hardness of SSV via 3-SAT Reduction}
\label{sec:np_hardness_proof}

First, we restate the SSV problem formulation. This is the same as the formalization in the main body of the paper, available here again for reference.

\begin{problem}[\problemname{Semantic Self-Verification (SSV)}]
    \leavevmode
    \begin{itemize}
        \item \textbf{Input:}
            \begin{enumerate}
                \item A statement \(S\).
                \item A semantic framework \(F\).
            \end{enumerate}
        \item \textbf{Output:} TRUE if statement \(S\) accurately describes its own semantic properties under framework \(F\); FALSE otherwise.
    \end{itemize}
\end{problem}

\subsection{Components of the Semantic Framework {\(F\)}}
\label{ssec:ssv_framework_components}

To make SSV precise, we define what a "semantic framework" consists of. This framework provides all the necessary components for interpreting the statement \(S\) and verifying its claims about itself.

For the purpose of our reduction, we define the semantic framework \(F = (T, \Sigma, \text{Cons}, M, V)\) where:
\begin{itemize}
    \item \(T = \{t_1, t_2, \dots, t_n\}\): A finite set of \textit{ambiguous terms} present in or implicitly referred to by statement \(S\).
    
    These are words or phrases in statement \(S\) that could have multiple meanings.
    
    \item \(\Sigma = \{\sigma_{1}^0, \sigma_{1}^1, \sigma_{2}^0, \sigma_{2}^1, \dots, \sigma_{n}^0, \sigma_{n}^1\}\): A finite set of possible \textit{elemental meanings} or \textit{senses}. In our simplified model for the proof, each ambiguous term \(t_i \in T\) is associated with exactly two possible elemental meanings: \(\sigma_i^0\) and \(\sigma_i^1\).
    
    For each ambiguous term \(t_i\), we give it exactly two possible interpretations, \(\sigma_i^0\) or \(\sigma_i^1\). This binary choice will directly correspond to the TRUE/FALSE assignments in 3-SAT.
    
    \item \(\text{Cons} = \{SC_1, SC_2, \dots, SC_m\}\): A finite set of \textit{semantic constraints}. Each constraint \(SC_j\) is a condition on the meanings assigned to a subset of terms in \(T\).
    
    These are rules that dictate which combinations of meanings are allowed. For example, a constraint might say "If term \(t_1\) has meaning \(\sigma_1^0\), then term \(t_2\) cannot have meaning \(\sigma_2^1\)."
    
    \item \(M\): The \textit{Meaning Function}.
        \begin{itemize}
            \item An \textit{interpretation} \(I\) of statement \(S\) (with respect to its ambiguous terms in \(T\)) is a function \(I: T \to \bigcup_{i=1}^{n} \{\sigma_i^0, \sigma_i^1\}\) such that for each \(t_i \in T\), \(I(t_i) \in \{\sigma_i^0, \sigma_i^1\}\).
            
            An interpretation is simply a choice of one specific meaning (out of its two possibilities) for every ambiguous term in the statement.
            
            \item \(M(S, F)\) (or simply \(M(S)\) when F is clear) is the set of all \(2^n\) possible interpretations of \(S\).
            
            If there are \(n\) ambiguous terms, each with 2 meanings, there are \(2^n\) ways to interpret the statement as a whole.
            
        \end{itemize}
    \item \(V\): The \textit{Verification Function}.
        \begin{itemize}
            \item \(V(S, I, F)\) (or \(V(S, I)\)) = TRUE if interpretation \(I\) of \(S\) satisfies all semantic constraints in \(\text{Cons}\); FALSE otherwise.
            
            This function checks if a particular interpretation \(I\) is "valid" by seeing if it obeys all the rules (semantic constraints).
            
            \item The overall SSV problem then asks whether \(S\)'s specific self-referential claim about the existence of such a satisfying interpretation is true.
            
            Statement \(S\) itself will make a claim (e.g. "There exists an interpretation of my terms that satisfies all my constraints"). The SSV problem is about checking if that very claim made by \(S\) is correct according to the framework \(F\).
        \end{itemize}
\end{itemize}

We prove that SSV is NP-hard by showing that any instance of the 3-SAT problem can be transformed, in polynomial time, into an instance of our SSV problem. If we can do this transformation such that the original 3-SAT problem is solvable if and only if our new SSV problem is solvable, then SSV must be at least as hard as 3-SAT. This is denoted by \(3\text{-SAT} \le_p \text{SSV}\).

\subsection{Construction of the SSV Instance from a 3-SAT Formula \(\phi\)}
\label{ssec:construction}

Given a 3-SAT formula \(\phi\) with \(n\) variables \(x_1, \dots, x_n\) and \(m\) clauses \(C_1, \dots, C_m\). We construct an SSV instance \((S_{\phi}, F_{\phi})\) as follows:

We build a specific statement \(S_{\phi}\) and a specific semantic framework \(F_{\phi}\) from the given 3-SAT formula \(\phi\). The goal is to mirror the structure of \(\phi\) within our SSV components.

\begin{enumerate}
    \item \textbf{Construct the Set of Ambiguous Terms \(T_{\phi}\):} \\
    For each Boolean variable \(x_i\) in \(\phi\), create a corresponding ambiguous term \(t_i \in T_{\phi}\). So, \(T_{\phi} = \{t_1, \dots, t_n\}\).

    \item \textbf{Construct the Set of Elemental Meanings \(\Sigma_{\phi}\):} \\
    For each ambiguous term \(t_i\), define two elemental meanings:
    \begin{itemize}
        \item \(\sigma_i^0\): representing the assignment \(x_i = \text{FALSE}\).
        \item \(\sigma_i^1\): representing the assignment \(x_i = \text{TRUE}\).
    \end{itemize}
    
    The two possible truth values for a 3-SAT variable \(x_i\) become the two possible "meanings" for our term \(t_i\). \(\sigma_i^0\) means \(x_i\) is false, \(\sigma_i^1\) means \(x_i\) is true.

    \item \textbf{Construct the Set of Semantic Constraints \(\text{Cons}_{\phi}\):} \\
    For each clause \(C_j = (l_{j1} \lor l_{j2} \lor l_{j3})\) in \(\phi\), create a corresponding semantic constraint \(SC_j \in \text{Cons}_{\phi}\).
    The constraint \(SC_j\) is satisfied by an interpretation \(I\) if and only if at least one of the following conditions holds:
    \begin{itemize}
        \item If the first literal in \(C_j\) is \(l_{j1} = x_k\) (variable \(x_k\) is not negated), then interpretation \(I\) must assign meaning \(\sigma_k^1\) to term \(t_k\) (i.e., \(I(t_k) = \sigma_k^1\)).
        \item If the first literal in \(C_j\) is \(l_{j1} = \neg x_k\) (variable \(x_k\) is negated), then interpretation \(I\) must assign meaning \(\sigma_k^0\) to term \(t_k\) (i.e., \(I(t_k) = \sigma_k^0\)).
        \item (Similarly for the second literal \(l_{j2}\) in \(C_j\), which corresponds to some term \(t_p\)).
        \item (Similarly for the third literal \(l_{j3}\) in \(C_j\), which corresponds to some term \(t_q\)).
    \end{itemize}
    Essentially, \(SC_j\) is satisfied if interpretation \(I\) assigns meanings to \(t_k, t_p, t_q\) (the terms corresponding to the variables in clause \(C_j\)) that would make the original logical clause \(C_j\) true.
    
    Each clause in the 3-SAT formula becomes a "semantic rule" in our SSV setup. A semantic constraint \(SC_j\) is satisfied if the chosen meanings for its terms make the original 3-SAT clause \(C_j\) true. For instance, if a clause is \((x_1 \lor \neg x_2 \lor x_3)\), the corresponding semantic constraint would require that (meaning for \(t_1\) is \(\sigma_1^1\)) OR (meaning for \(t_2\) is \(\sigma_2^0\)) OR (meaning for \(t_3\) is \(\sigma_3^1\)).

    \item \textbf{Construct the Statement \(S_{\phi}\):} \\
    \(S_{\phi}\) is the following specific self-referential statement:
    \begin{quote}
    "This statement, \(S_{\phi}\), implicitly refers to a set of ambiguous terms \(T_{\phi} = \{t_1, \dots, t_n\}\) and is subject to a set of semantic constraints \(\text{Cons}_{\phi} = \{SC_1, \dots, SC_m\}\) (derived from a 3-SAT formula \(\phi\)). There exists at least one interpretation \(I \in M(S_{\phi}, F_{\phi})\) of these terms that simultaneously satisfies all semantic constraints in \(\text{Cons}_{\phi}\)."
    \end{quote}
    
    This statement \(S_{\phi}\) is carefully crafted. It explicitly says that it's about its own terms \(T_{\phi}\) and constraints \(\text{Cons}_{\phi}\). Critically, it makes an existential claim: it claims that there is a way to interpret its terms such that all its constraints are met. This claim directly mirrors the question 3-SAT asks: "Does there exist a truth assignment that satisfies all clauses?"

    \item \textbf{Define the Semantic Framework \(F_{\phi}\):} \\
    \(F_{\phi} = (T_{\phi}, \Sigma_{\phi}, \text{Cons}_{\phi}, M_{\phi}, V_{\phi})\), where:
    \begin{itemize}
        \item \(T_{\phi}, \Sigma_{\phi}, \text{Cons}_{\phi}\) are as constructed in steps 1-3 above.
        \item \(M_{\phi}(S_{\phi}, F_{\phi})\) yields the set of all \(2^n\) interpretations by assigning either \(\sigma_i^0\) or \(\sigma_i^1\) to each term \(t_i\).
        \item \(V_{\phi}(S_{\phi}, I, F_{\phi})\) = TRUE if interpretation \(I\) satisfies all semantic constraints \(SC_j \in \text{Cons}_{\phi}\); FALSE otherwise.
        \item The SSV problem instance \((S_{\phi}, F_{\phi})\) outputs TRUE if the claim made by statement \(S_{\phi}\) (from step 4) is true. That is, the SSV instance is TRUE if there indeed exists an interpretation \(I \in M_{\phi}(S_{\phi}, F_{\phi})\) such that \(V_{\phi}(S_{\phi}, I, F_{\phi}) = \text{TRUE}\). Otherwise, it outputs FALSE.
    \end{itemize}
    
    The framework \(F_{\phi}\) packages up all the pieces we just constructed. The core of the SSV problem here is: is the claim made inside \(S_{\phi}\) actually true given these rules?
    
\end{enumerate}

\subsection{Polynomial-Time Transformation}
\label{ssec:poly_time_transformation}

For a reduction to be valid in establishing NP-hardness, the transformation process itself must be efficient; it shouldn't be harder than the problem we're trying to solve.

The construction of the SSV instance \((S_{\phi}, F_{\phi})\) from the 3-SAT formula \(\phi\) can be performed in polynomial time with respect to the size of \(\phi\) (which is determined by \(n\), the number of variables, and \(m\), the number of clauses).

\begin{itemize}
    \item Constructing \(T_{\phi}\) (terms from variables) takes time proportional to \(n\), denoted \(O(n)\).
    \item Constructing \(\Sigma_{\phi}\) (meanings for terms) also takes \(O(n)\) time.
    \item Constructing \(\text{Cons}_{\phi}\): For each of \(m\) clauses in \(\phi\), creating the corresponding semantic constraint \(SC_j\) involves looking at 3 literals. This description can be generated in constant time per clause. So, total time is \(O(m)\).
    \item Constructing statement \(S_{\phi}\): The textual representation of \(S_{\phi}\) can be written down such that it lists or symbolically refers to these \(n\) terms and \(m\) constraints. Its length will be polynomial in \(n+m\).
    \item The framework \(F_{\phi}\) is simply defined by these polynomially-sized components.
\end{itemize}
Thus, the overall transformation from \(\phi\) to \((S_{\phi}, F_{\phi})\) is achieved in polynomial time.

\subsection{Proof of Equivalence (Correctness)}
\label{ssec:proof_of_equivalence}

Here, we show that our constructed SSV instance \((S_{\phi}, F_{\phi})\) gives a TRUE output if and only if the original 3-SAT formula \(\phi\) is satisfiable. We prove this in two parts.

\textbf{Forward Direction: If \(\phi\) is satisfiable, then SSV instance \((S_{\phi}, F_{\phi})\) is TRUE.}

First, we assume we have a solution to the 3-SAT problem (\(\phi\) is satisfiable) and show that this implies our SSV problem also has a "yes" answer (i.e., statement \(S_{\phi}\) is true).

\begin{proof}
    \begin{enumerate}
        \item Assume \(\phi\) is satisfiable. This means there exists a truth assignment \(A^* = \{a_1^*, \dots, a_n^*\}\) (where each \(a_i^*\) is either TRUE or FALSE) that makes the entire formula \(\phi\) true.
        \item We construct an interpretation \(I^*\) for the terms in \(T_{\phi}\) based on this satisfying assignment \(A^*\): For each term \(t_i \in T_{\phi}\), if \(a_i^* = \text{TRUE}\), we set the meaning \(I^*(t_i) = \sigma_i^1\). If \(a_i^* = \text{FALSE}\), we set \(I^*(t_i) = \sigma_i^0\).
        \item Consider an arbitrary clause \(C_j = (l_{j1} \lor l_{j2} \lor l_{j3})\) in \(\phi\). Since the assignment \(A^*\) satisfies \(\phi\), it must satisfy every clause \(C_j\). This means at least one of its three literals is made true by \(A^*\). Let this true literal be \(l_{jk}\) for some \(k \in \{1, 2, 3\}\).
        \item We show that because \(A^*\) satisfies \(C_j\), our constructed interpretation \(I^*\) must satisfy the corresponding semantic constraint \(SC_j\). We analyze the two possible forms of the literal \(l_{jk}\):
            \begin{itemize}
                \item \textbf{Case 1: The literal is positive.} Let \(l_{jk} = x_i\) for some variable \(x_i\). For \(A^*\) to make \(x_i\) true, we must have \(a_i^* = \text{TRUE}\). By our construction of \(I^*\) in step 2, \(a_i^* = \text{TRUE}\) implies that \(I^*(t_i) = \sigma_i^1\). By our construction of the semantic constraint \(SC_j\), an interpretation satisfies \(SC_j\) if the meaning assigned to the term corresponding to \(x_i\) (which is \(t_i\)) is \(\sigma_i^1\). Since \(I^*(t_i) = \sigma_i^1\), \(I^*\) satisfies \(SC_j\).
                \item \textbf{Case 2: The literal is negative.} Let \(l_{jk} = \neg x_i\) for some variable \(x_i\). For \(A^*\) to make \(\neg x_i\) true, we must have \(a_i^* = \text{FALSE}\). By our construction of \(I^*\) in step 2, \(a_i^* = \text{FALSE}\) implies that \(I^*(t_i) = \sigma_i^0\). By our construction of \(SC_j\), an interpretation satisfies \(SC_j\) if the meaning assigned to the term corresponding to \(\neg x_i\) (which is \(t_i\)) is \(\sigma_i^0\). Since \(I^*(t_i) = \sigma_i^0\), \(I^*\) satisfies \(SC_j\).
            \end{itemize}
            In both cases, if a literal in \(C_j\) is true under \(A^*\), the corresponding condition in \(SC_j\) is met by \(I^*\). Since at least one such literal must be true, \(I^*\) is guaranteed to satisfy \(SC_j\).
        \item The logic in step 4 holds for any arbitrary clause \(C_j\). Since our chosen interpretation \(I^*\) satisfies \textit{all} semantic constraints \(SC_j \in \text{Cons}_{\phi}\) (because \(A^*\) satisfied all clauses \(C_j\)), the verification function \(V_{\phi}(S_{\phi}, I^*, F_{\phi})\) will output TRUE.
        \item Statement \(S_{\phi}\) (from step 4 in Section \ref{ssec:construction}) makes the claim: "There exists at least one interpretation ... that simultaneously satisfies all semantic constraints." Because we have just found such an interpretation (\(I^*\)), the claim made by \(S_{\phi}\) is true.
        \item Therefore, by the definition of our SSV problem, the SSV instance \((S_{\phi}, F_{\phi})\) outputs TRUE.
    \end{enumerate}
\end{proof}

\textbf{Backward Direction: If SSV instance \((S_{\phi}, F_{\phi})\) is TRUE, then \(\phi\) is satisfiable.}

Now, we prove the other direction. We assume our SSV problem has a "yes" answer (statement \(S_{\phi}\) is true) and show that this implies the original 3-SAT problem also has a solution (\(\phi\) is satisfiable).

\begin{proof}
    \begin{enumerate}
        \item Assume the SSV instance \((S_{\phi}, F_{\phi})\) is TRUE. According to the definition of SSV (Problem \ref{ssec:ssv_problem_statement}), this means the statement \(S_{\phi}\) is evaluated as accurately describing its own semantic properties under the framework \(F_{\phi}\).
        \item The statement \(S_{\phi}\) makes the specific claim: "There exists at least one interpretation \(I \in M(S_{\phi}, F_{\phi})\) ... that simultaneously satisfies all semantic constraints in \(\text{Cons}_{\phi}\)." Since \(S_{\phi}\) is true, this claim must be true. Therefore, there must exist at least one such interpretation, which we will call \(I^{**}\), for which \(V_{\phi}(S_{\phi}, I^{**}, F_{\phi}) = \text{TRUE}\).
        \item This means that the interpretation \(I^{**}\) satisfies all semantic constraints \(SC_j \in \text{Cons}_{\phi}\).
        \item We construct a truth assignment \(A^{**} = \{a_1^{**}, \dots, a_n^{**}\}\) for the variables of \(\phi\) based on \(I^{**}\): For each variable \(x_i\), if \(I^{**}(t_i) = \sigma_i^1\), we set \(a_i^{**} = \text{TRUE}\). If \(I^{**}(t_i) = \sigma_i^0\), we set \(a_i^{**} = \text{FALSE}\).
        \item Consider an arbitrary semantic constraint \(SC_j\) corresponding to clause \(C_j = (l_{j1} \lor l_{j2} \lor l_{j3})\). Since \(I^{**}\) satisfies \(SC_j\), at least one of its three disjunctive conditions must be met. Let this condition correspond to the literal \(l_{jk}\).
        \item We now show that because \(I^{**}\) satisfies \(SC_j\), our constructed assignment \(A^{**}\) must satisfy the corresponding clause \(C_j\). We analyze the two possible forms of the condition met by \(I^{**}\):
            \begin{itemize}
                \item \textbf{Case 1: The condition corresponds to a positive literal \(x_i\).} For this condition to be met, we must have \(I^{**}(t_i) = \sigma_i^1\). By our construction of \(A^{**}\) in step 4, this implies that \(a_i^{**} = \text{TRUE}\). This assignment makes the literal \(x_i\) true. Since \(x_i\) is one of the literals in the disjunction \(C_j\), the clause \(C_j\) is satisfied by \(A^{**}\).
                \item \textbf{Case 2: The condition corresponds to a negative literal \(\neg x_i\).} For this condition to be met, we must have \(I^{**}(t_i) = \sigma_i^0\). By our construction of \(A^{**}\) in step 4, this implies that \(a_i^{**} = \text{FALSE}\). This assignment makes the literal \(\neg x_i\) true. Since \(\neg x_i\) is one of the literals in the disjunction \(C_j\), the clause \(C_j\) is satisfied by \(A^{**}\).
            \end{itemize}
            In both cases, a condition in \(SC_j\) being met by \(I^{**}\) guarantees that the corresponding literal in \(C_j\) is made true by \(A^{**}\).
        \item The logic in step 6 holds for any arbitrary constraint \(SC_j\). Because \(I^{**}\) satisfies \textit{all} semantic constraints, the constructed truth assignment \(A^{**}\) must therefore satisfy \textit{all} corresponding clauses \(C_j\) in the formula \(\phi\).
        \item Therefore, the 3-SAT formula \(\phi\) is satisfiable (using the assignment \(A^{**}\)).
    \end{enumerate}
\end{proof}

Since we have shown that \(\phi\) is satisfiable if and only if \((S_{\phi}, F_{\phi})\) is TRUE, and the transformation is polynomial-time, we establish that \(3\text{-SAT} \le_p \text{SSV}\). Given that 3-SAT is NP-complete, this proves that SSV is at least NP-hard.

\section{SSV is in NP}
\label{sec:ssv_in_np}

To show that SSV is NP-complete, we also need to show that this specification of the problem belongs to the class NP. A problem is in NP if a proposed solution (a "certificate" for a "yes" instance) can be checked for correctness efficiently (in polynomial time).

A decision problem is in the class NP if for any "yes" instance, there exists a certificate (or proof) that can be verified in polynomial time by a deterministic algorithm.

\paragraph{Certificate for SSV:} For an SSV instance \((S, F)\) where the answer is TRUE (meaning statement \(S\)'s self-referential claim about satisfying its constraints is true), a certificate is an actual interpretation \(I \in M(S,F)\) that indeed satisfies all semantic constraints in \(\text{Cons}\) (as claimed by \(S\)).

If \(S\) claims "there exists a good interpretation," a "good interpretation" itself is the proof.
    
\paragraph{Verification Algorithm:} Given the SSV instance \((S_{\phi}, F_{\phi})\) (whose description size is polynomial in the original 3-SAT formula \(\phi\)) and a proposed certificate interpretation \(I\):
    \begin{enumerate}
        \item Check if \(I\) is a valid interpretation: For each of the \(n\) ambiguous terms \(t_i \in T_{\phi}\), ensure that \(I(t_i)\) is one of its two allowed elemental meanings (\(\sigma_i^0\) or \(\sigma_i^1\)). This check takes \(O(n)\) time.
        \item Verify constraints: For each of the \(m\) semantic constraints \(SC_j \in \text{Cons}_{\phi}\), check if the given interpretation \(I\) satisfies \(SC_j\). Since each \(SC_j\) is derived from a 3-literal clause in 3-SAT, it refers to the meanings of at most 3 terms. Checking one such constraint \(SC_j\) against the interpretation \(I\) takes constant time (or time related to looking up 3 term meanings). Verifying all \(m\) constraints thus takes \(O(m)\) time.
    \end{enumerate}
The total time for this verification process is \(O(n+m)\), which is polynomial in the size of the input SSV instance (which itself is polynomial in the size of the original \(\phi\)).

Therefore, since a "yes" answer to SSV can be verified in polynomial time given a suitable certificate, SSV is in NP.

\section{Conclusion of Complexity Argument}
\label{sec:complexity_conclusion}

Since SSV is NP-hard and SSV is in NP, \textbf{SSV is NP-complete.}

NP-complete problems are the "hardest" problems in NP. They are all equivalent in difficulty (up to polynomial-time transformations). For practical purposes, NP-completeness signals that we should not expect efficient and exact algorithms for all instances of SSV as we've defined it.

\end{document}